\documentclass[a4paper,fleqn]{caswenjia}

\usepackage[numbers]{natbib}
\usepackage{geometry}
\usepackage{amsmath}
\usepackage{amsfonts}
\usepackage{amssymb}
\usepackage{bbm}
\usepackage{hyperref}
\usepackage{verbatim}
\usepackage{caption}
\usepackage{adjustbox}
\usepackage{url}
\usepackage{threeparttable}
\usepackage[normalem]{ulem}
\usepackage{color}
\usepackage[inline]{enumitem}
\usepackage{xcolor}
\usepackage{colortbl}

\usepackage{makecell} 
\usepackage{pifont}   
\usepackage{adjustbox} 

\newcommand{\CheckmarkBold}{\ding{51}}
\newcommand{\XSolidBrush}{\ding{55}}
\newcommand{\header}[1]{\textbf{\makecell{#1}}}

\definecolor{lightgreen}{rgb}{0, 0.69, 0.31}
\definecolor{lightblue}{rgb}{0, 0.44, 0.75}
\definecolor{brightblue}{rgb}{0, 0.69, 0.94}
\definecolor{deepred}{rgb}{0.75, 0, 0}
\definecolor{graybg}{gray}{0.95}

\newcommand{\etal}{\textit{et al.}}

\newcommand{\myparagraph}[1]{\vspace{2pt}\noindent{\bf #1}}

\def\tsc#1{\csdef{#1}{\textsc{\lowercase{#1}}\xspace}}
\tsc{WGM}
\tsc{QE}
\tsc{EP}
\tsc{PMS}
\tsc{BEC}
\tsc{DE}

\begin{document}
\begin{sloppypar}

\let\WriteBookmarks\relax
\def\floatpagepagefraction{1}
\def\textpagefraction{.001}

\shorttitle{UAV traffic scene understanding: A regulation embedded multi-modal network and a unified benchmark}

\shortauthors{Yu Zhang et~al.}




\title[mode = title]{UAV traffic scene understanding: A regulation embedded multi-modal network and a unified benchmark}

\author[1,2]{Yu Zhang}

\author[3,4]{Zhicheng Zhao}
\cormark[1]
\ead{zhaozhicheng@ahu.edu.cn}
\author[1,2]{Ze Luo}
\cormark[1]
\ead{luoze@cnic.cn}
\author[3,4]{Chenglong Li}
\author[3,4]{Jin Tang}

\address[1]{Computer Network Information Center, Chinese Academy of Sciences, Beijing 100190, China}
\address[2]{University of Chinese Academy of Sciences, Beijing 100190, China}
\address[3]{Anhui Provincial Key Laboratory of Multi-modal Cognitive Computation, Anhui University, Hefei 230601, China}
\address[4]{Information Materials and Intelligent Sensing Laboratory of Anhui Province, Anhui University, Hefei 230601, China}

\begin{keywords}
Visual Question Answering\sep 
UAV Traffic Scene Understanding\sep
Optical-Thermal\sep 
Multi-Modal Fusion\sep
Cognitive Reasoning\sep 

\end{keywords}
\let\printorcid\relax

\maketitle

\begin{abstract}
Traffic scene understanding from unmanned aerial vehicle (UAV) platforms is crucial for intelligent transportation systems due to its flexible deployment and wide-area monitoring capabilities. However, existing methods face significant challenges in real-world surveillance, as their heavy reliance on optical imagery leads to severe performance degradation under adverse illumination conditions like nighttime and fog. Furthermore, current Visual Question Answering (VQA) models are restricted to elementary perception tasks, lacking the domain-specific regulatory knowledge required to assess complex traffic behaviors. To address these limitations, we propose a novel Multi-modal Traffic Cognition Network (MTCNet) for robust UAV traffic scene understanding. Specifically, we design a Prototype-Guided Knowledge Embedding (PGKE) module that leverages high-level semantic prototypes from an external Traffic Regulation Memory (TRM) to anchor domain-specific knowledge into visual representations, enabling the model to comprehend complex behaviors and distinguish fine-grained traffic violations. Moreover, we develop a Quality-Aware Spectral Compensation (QASC) module that exploits the complementary characteristics of optical and thermal modalities to perform bidirectional context exchange, effectively compensating for degraded features to ensure robust representation in complex environments. In addition, we construct Traffic-VQA, the first large-scale optical-thermal infrared benchmark for cognitive UAV traffic understanding, comprising 8,180 aligned image pairs and 1.3 million question-answer pairs across 31 diverse types. Extensive experiments demonstrate that MTCNet significantly outperforms state-of-the-art methods in both cognition and perception scenarios. The dataset is available at \url{https://github.com/YuZhang-2004/UAV-traffic-scene-understanding}.

\end{abstract}

\section{Introduction}

\begin{figure*}
  \centering
  \includegraphics[width=1.0\textwidth]{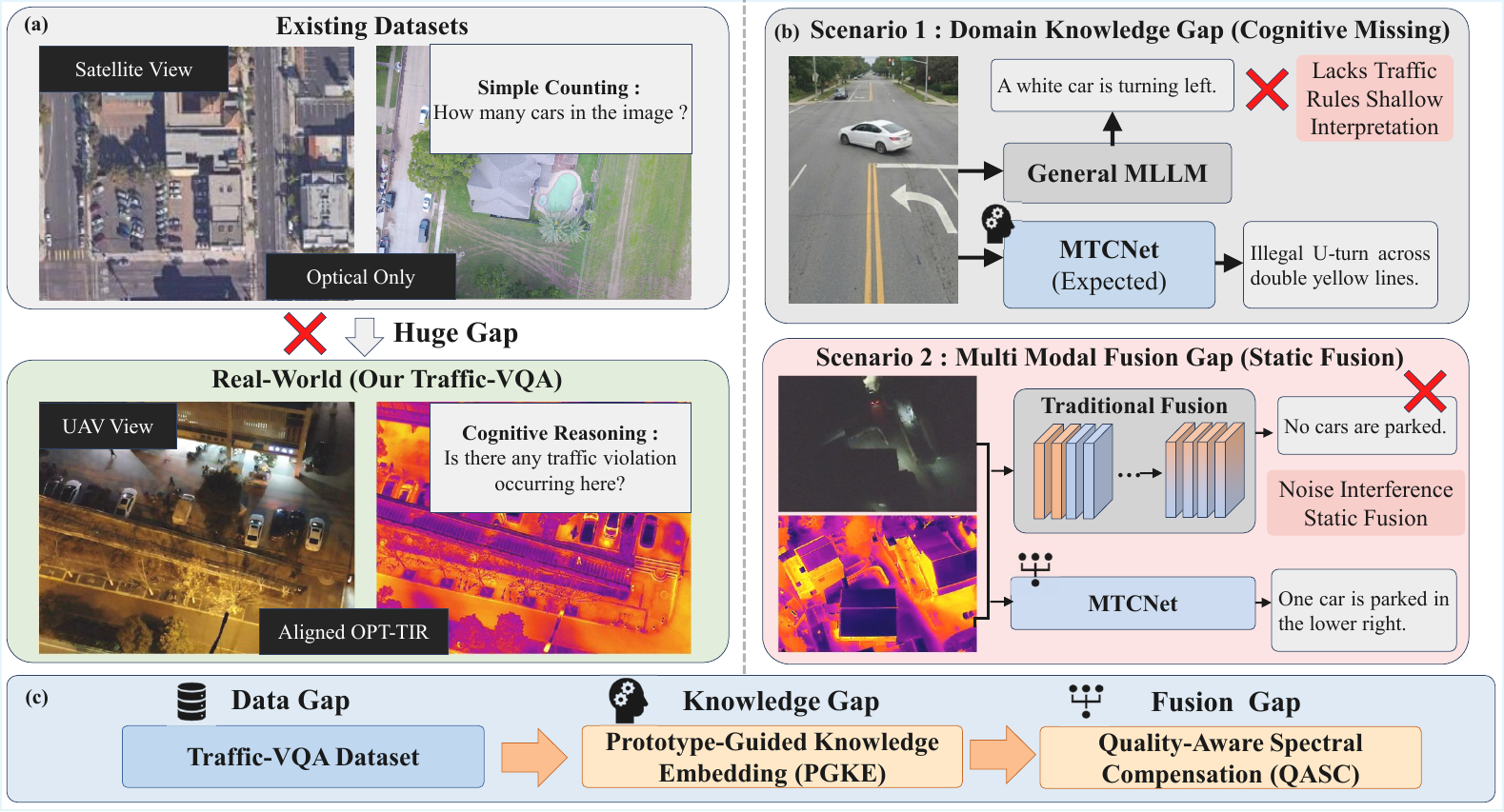}
  \caption{Current challenges in UAV-based traffic VQA. (a) Data Gap. Existing datasets (top) rely on single-modal optical imagery for elementary perception, whereas practical surveillance (bottom) demands aligned OPT-TIR data for complex cognitive understanding. (b) Methodological Bottlenecks. General MLLMs struggle with the \textit{Domain Knowledge Gap}, failing to interpret specific traffic rules (\textit{e.g.}, missing the "illegal" attribute of a U-turn), and the \textit{Multi-Modal Fusion Gap}, where static fusion allows degraded optical noise to corrupt robust thermal features under adverse conditions. (c) Our Solution. The proposed MTCNet systematically bridges these gaps through the Traffic-VQA dataset, the PGKE module, and the QASC module.}
  \label{fig:motivation}
\end{figure*}

With the continuous advancement of remote sensing observation technology, Unmanned Aerial Vehicles (UAVs) have become essential components of Intelligent Transportation Systems (ITS). Compared to fixed ground-level surveillance cameras, UAVs offer flexible deployment and wide-area monitoring capabilities, enabling comprehensive observation of dynamic traffic flows. Consequently, UAV-based Visual Question Answering (UAV-VQA)~\cite{VQA} has emerged as a transformative technology. Unlike conventional visual approaches constrained to fundamental object detection and counting, UAV-VQA enables operators to interact with traffic scenes through natural language queries (\textit{e.g.}, "\textit{Is the white vehicle at the intersection violating traffic rules?}"), thereby providing actionable insights for dynamic traffic control and safety regulation.

However, deploying UAV-VQA in real-world traffic surveillance presents significant scientific challenges. Beyond controlled ideal laboratory settings, practical UAV imagery is captured under unconstrained open-world conditions. These scenes are characterized by extreme viewpoint variations, dense distributions of tiny objects (\textit{e.g.}, miniaturized vehicles in wide-angle views), and severe vulnerability to environmental degradation such as nighttime darkness, glare, or dense fog~\cite{Guidance_disentanglement}. Furthermore, interpreting complex traffic behaviors requires domain-specific regulatory knowledge, exposing a critical cognitive deficiency in current visual systems. As illustrated in Fig. \ref{fig:motivation}, these factors lead to substantial visual feature corruption and semantic ambiguity, severely bottlenecking the perception capabilities of traditional visual models and limiting the practical deployment of UAV-VQA systems.

While existing VQA methods and Multi-modal Large Language Models (MLLMs) designed for aerial platforms have driven significant progress in generalized spatial comprehension, they exhibit notable limitations. Representative methods such as GeoChat~\cite{geochat} and EarthGPT~\cite{earthgpt,EarthGPT-X} leverage pre-trained large models as unified interfaces to handle diverse interpretation tasks. However, a major vulnerability of these approaches is their predominant reliance on high-quality, single-modality optical (OPT) imagery. Since extracted feature representations are highly sensitive to input quality, optical signals are easily disrupted by environmental noise~\cite{Data_fusion}. Consequently, these models often suffer severe perception failures when applied to challenging, all-weather traffic scenarios where the optical signal is significantly degraded.

To mitigate the limitations of optical sensors, thermal infrared (TIR) imagery provides a complementary modality that captures heat signatures independent of ambient illumination, enabling effective performance in darkness and fog~\cite{gujuanjuan, Guidance_disentanglement, liu2024cross, zhao2025reflectance}. Fusing optical and TIR data represents a highly promising strategy for robust, all-weather perception. Mainstream multi-modal fusion strategies have evolved from simple static concatenation to sophisticated dynamic attention and correlation-driven mechanisms. Despite their success in low-level perception tasks such as object detection, these advanced multi-modal fusion techniques have rarely been integrated into the high-level cognitive context of UAV-VQA, leaving a significant gap in robust multi-modal feature representation for traffic behavior understanding.

Regarding cognitive capabilities, state-of-the-art MLLMs have made remarkable strides in generalized visual reasoning and complex deductive tasks~\cite{gemini2.5pro,qwen2.5-vl,deepseek-vl,gpt4o}. However, when deployed in the highly dynamic and regulation-intensive traffic scenarios typical of UAV traffic surveillance, these models encounter a fundamental cognitive bottleneck. Current MLLMs predominantly rely on broad statistical priors and surface-level visual-semantic alignment, critically lacking the capacity to ground complex, evolving visual states into domain-specific regulatory frameworks. For instance, as illustrated in Fig.~\ref{fig:motivation}, when a vehicle crosses double yellow lines, a general-purpose MLLM may provide only a shallow spatial description (\textit{e.g.}, "A white car is turning left"), fundamentally missing the "illegal" attribute of the U-turn. This failure occurs because the model cannot map the dynamic trajectory to the abstract traffic rule. Without explicit regulatory memory to anchor such specialized rules, these models struggle to maintain logical consistency in unconstrained traffic environments, often resorting to spurious visual correlations that result in semantic hallucinations and a critical perception-cognition disconnect.

To address these combined challenges of multi-spectral data fusion and domain-specific cognitive depth, we propose a novel Multi-modal Traffic Cognition Network (MTCNet). We decouple the problem into two complementary objectives: enhancing low-level perceptual robustness and injecting high-level cognitive context. Specifically, we design a Prototype-Guided Knowledge Embedding (PGKE) module that leverages an external Traffic Regulation Memory (TRM) constructed from expert knowledge. By retrieving and aligning high-level semantic prototypes with current visual features, the PGKE module injects domain-specific cognitive capabilities into the network. Simultaneously, we develop a Quality-Aware Spectral Compensation (QASC) module that orchestrates a bidirectional context exchange between optical and thermal modalities via a dynamic attention mechanism, allowing the network to selectively transfer discriminative features from the reliable modality to compensate for the degraded one, ensuring robust representation in complex environments. Beyond these methodological contributions, the research field also lacks a large-scale benchmark dataset with aligned multi-spectral imagery and cognitive annotations, which is crucial for training and evaluating robust traffic understanding models. To fill this gap, we construct Traffic-VQA, a large-scale OPT-TIR benchmark for UAV traffic understanding. It contains 8,180 meticulously aligned image pairs with over 1.3 million question-answer pairs, covering diverse environmental conditions (sunny, night, fog) and a comprehensive taxonomy of 31 question types, spanning from basic perception to complex violation understanding. Extensive experiments conducted on the proposed dataset demonstrate the superiority and effectiveness of MTCNet compared to existing state-of-the-art methods. The main contributions of this work are summarized as follows:
\begin{itemize}

    \item We introduce Traffic-VQA, the first large-scale OPT-TIR benchmark dedicated to cognitive traffic understanding. This dataset contains a substantial number of complex traffic scenarios and cognition-oriented QA pairs, providing a critical foundation for advancing all-weather UAV perception and cognition tasks.

    \item Recognizing the vulnerability of single optical signals and the lack of domain-specific cognitive depth in current UAV traffic scene understanding, we propose the MTCNet framework to effectively harness complementary optical-thermal features and integrate domain-specific regulatory knowledge.

    \item To bridge the domain knowledge gap, we design a PGKE module, which retrieves and embeds domain-specific regulatory semantics via a constructed traffic regulation memory bank. Furthermore, to enhance low-level perceptual robustness under adverse conditions, we further introduce a QASC module to guide the network in performing dynamic bidirectional context exchange between optical and thermal modalities.

    \item Extensive experiments on the Traffic-VQA dataset validate the effectiveness of the proposed MTCNet. Compared with existing leading open-source MLLMs (\textit{e.g.}, Qwen2.5-VL~\cite{qwen2.5-vl}, GeoChat~\cite{geochat}) and closed-source commercial models (\textit{e.g.}, GPT-4o~\cite{gpt4o}), our method successfully anchors domain-specific regulatory knowledge, achieving substantial improvements in comprehension accuracy for complex traffic behaviors across all-weather scenarios.

\end{itemize}

\section{Related Work}

The proposed MTCNet framework builds upon and advances three interrelated research areas: visual question answering, multi-modal integration, and the cognitive capabilities of large language models. In what follows, we review representative works in each area and discuss their limitations in the context of UAV traffic understanding.

\subsection{Visual Question Answering}
VQA for aerial imagery has evolved significantly, progressing from modular specialist architectures to integrated analytical systems. Early methods predominantly utilized dual-stream architectures, employing Convolutional Neural Networks (CNNs) for visual feature extraction and Recurrent Neural Networks (RNNs) for question encoding \cite{RSVQA, rsivqa}. These foundational models, alongside early large-scale benchmarks such as RSVQA-LR and RSVQA-HR \cite{RSVQA}, typically aggregated features via basic element-wise operations. However, such approaches often struggled to capture complex geospatial interactions \cite{zhang2023spatial}. The scope of these initial methods was broadened by datasets like RSVQAxBEN \cite{RSVQA_Meets}, which introduced logical connectors into queries to incrementally increase the complexity of structural linguistics.

To address the misalignment between spatial visual layouts and linguistic tokens, subsequent research introduced refined attention mechanisms. For instance, MAIN \cite{rsivqa} leveraged mutual attention for bidirectional guidance, whereas SHRNet \cite{zhang2023spatial} and MQVQA \cite{MQVQA} incorporated spatial hierarchical reasoning and multistep attention to model high-order intra-group object relations \cite{zhang2023spatial, MQVQA}. As the field progressed toward practical deployments, researchers developed domain-specific frameworks. FloodNet \cite{floodnet} provided evaluations exclusively designed for post-disaster damage assessment, while CDVQA \cite{CDVQA} introduced multitemporal analytical reasoning to address semantic change detection. Furthermore, RSIVQA \cite{rsivqa} and TextRS \cite{VQA-TextRS} integrated multiple external sources to curate human-annotated answers that more closely mirror naturalistic human queries.

Despite these successes in fundamental perception tasks, early VQA methods were constrained by rigid answer spaces and limited scalability across varied interpretation tasks. They primarily focused on basic perceptual queries such as object presence and spatial counting, lacking the cognitive capacity to address complex, open-ended interpretation tasks in diverse operational environments.

\subsection{Multi-modal VQA}
Multi-modal VQA aims to leverage the complementary physics of heterogeneous sensors---such as optical and thermal or synthetic aperture radar (SAR) configurations---to increase informational density and analytical precision across diverse environmental contexts \cite{earthgpt, soni2025earthdial}. Within this area, researchers have systematically addressed representational discrepancies originating from distinct imaging modalities and environmental degradation. To mitigate basic modality disparities, Li \etal~\cite{MFFENet} introduced a multiscale feature fusion and enhancement network designed to amplify foreground semantic objects, thereby improving the parsing of urban road networks under suboptimal illumination. To address the intrinsic modality gap, Zhou \etal~\cite{M-SpecGene} developed M-SpecGene, a generalized foundation model that uses a cross-modality structural sparsity metric to quantify information density and extract modality-invariant representations. Concurrently, Zhao \etal~\cite{CDDFuse} proposed CDDFuse, a correlation-driven architecture that decomposes features into modality-shared and modality-specific sub-components to improve cross-modal consistency.

Moving beyond static integration approaches, Zhang \etal~\cite{M2FNet} formulated M2FNet, which dynamically aggregates multi-spectral features via union-modal and cross-modal attention mechanisms to ensure robust object detection regardless of illumination variance. Furthermore, Xu \etal~\cite{U2Fusion} proposed a unified unsupervised framework, U2Fusion, which autonomously estimates the informational saliency of source imagery to guide adaptive feature preservation. Generative and prompt-guided strategies have also shown promise in bridging representational gaps. Advanced architectures such as AT-GAN \cite{AT-GAN} and denoising diffusion models \cite{DDFM} have been adapted to synthesize high-fidelity fused imagery while preserving intricate structural details. Specifically targeting unconstrained aerial environments, Zhao \etal~\cite{Guidance_disentanglement} introduced GDNet for the disentanglement of optical guidance features, while PromptFusion \cite{PromptFusion} harmonized multi-modality images guided by explicit semantic prompts through bi-level optimization.

Despite these promising results across various applications, the potential of advanced integration techniques within explicit cognitive contexts remains largely unexplored. Most existing multi-modal methods rely on rigid integration strategies that lack the dynamic, non-destructive context exchange required to handle the extreme variance characteristic of all-weather traffic scenarios.

\subsection{Multi-modal Large Language Models}
Recent developments in MLLMs for aerial scenarios have transitioned toward open-set generalization by adopting pre-trained Large Language Models (LLMs) as unified cognitive interfaces. Early efforts, such as EarthGPT~\cite{earthgpt} and EarthGPT-X~\cite{EarthGPT-X}, integrated various multi-sensor interpretation tasks through cross-modal comprehension and visual prompting. To incorporate spatial-temporal and geo-context clues, SkySense~\cite{skysense} introduced a factorized multi-modal spatiotemporal encoder pre-trained on a large scale. Similarly, RingMoGPT~\cite{RingMoGPT} unified vision, language, and localization tasks using a location- and instruction-aware querying transformer. While these foundational architectures established versatile baselines for multi-sensor data, their holistic scene interpretation often lacked the fine-grained spatial awareness required for small, densely distributed objects.

To address unique spatial complexities, subsequent research focused on region-level reasoning and high-resolution processing. GeoChat~\cite{geochat} advanced local perception by accepting region inputs for region-specific dialogues and visual grounding. Extending this to dynamic scenarios, EarthDial~\cite{soni2025earthdial} enabled multi-sensory interactive dialogues, while AirSpatialBot~\cite{airspatialbot} specifically targeted fine-grained vehicle attribute recognition and retrieval using a 3D visual grounding approach. Furthermore, handling ultra-high-resolution imagery poses significant token explosion challenges. To address this, GeoLLaVA-8K~\cite{geollava8k} utilized background token pruning, and LRS-VQA~\cite{When_Large} proposed a coarse-to-fine text-guided token pruning strategy. Despite improving spatial grounding, these methods predominantly rely on static feature extraction pathways that struggle with complex deductive logic.

To further enhance interpretive depth, inference-centric approaches utilizing reinforcement learning have recently gained traction. Frameworks such as Geo-R1~\cite{geo-r1} and Vision-R1~\cite{vision-r1} utilize verifiable rewards and group relative policy optimization to incentivize genuine geospatial reasoning. To overcome human annotation biases, GeoZero~\cite{geozero} attempts to elicit reasoning from scratch without predefined templates, while RS-EoT~\cite{RS-EoT} employs an iterative evidence-seeking approach to mitigate pseudo-reasoning and the glance effect. However, comprehensive evaluations indicate that these models still struggle with dense, complex imagery and remain susceptible to semantic hallucinations during highly specialized interpretation tasks.

Despite these notable advancements, existing MLLMs exhibit a critical perception-cognition gap when deployed in specialized domains like UAV traffic surveillance. While proficient at detecting general entities, they consistently fail to decode complex traffic violations that require implicit regulatory knowledge. Furthermore, the absence of robust mechanisms to align multi-spectral observations with expert situational memory underscores the need for a novel architecture that explicitly integrates specialized domain rules into the visual interpretation process.

\section{Methodology}
\label{sec:method}

In this section, we present the proposed MTCNet framework, a cognitive prototype-anchored architecture designed to augment MLLMs for robust traffic reasoning from UAV imagery. As illustrated in Fig.~\ref{fig:overall}, our approach addresses two fundamental challenges in RSVQA: the domain knowledge gap inherent in interpreting specialized traffic behaviors, and the robust fusion of heterogeneous sensor modalities (OPT and TIR).

Specifically, we first describe the overall Gated Parallel Residual Architecture (Section~\ref{sec:overall}). We then detail the construction of the offline TRM (Section~\ref{sec:cpb}), which serves as an external knowledge repository. Subsequently, we elaborate on the two core trainable modules: the PGKE module (Section~\ref{sec:pgke}) for explicit domain knowledge injection, and the QASC module (Section~\ref{sec:qasc}) for dynamic, context-aware multi-spectral integration.

\subsection{Overall Architecture}
\label{sec:overall}

\begin{figure*}
\centering
\includegraphics[width=1.0\textwidth]{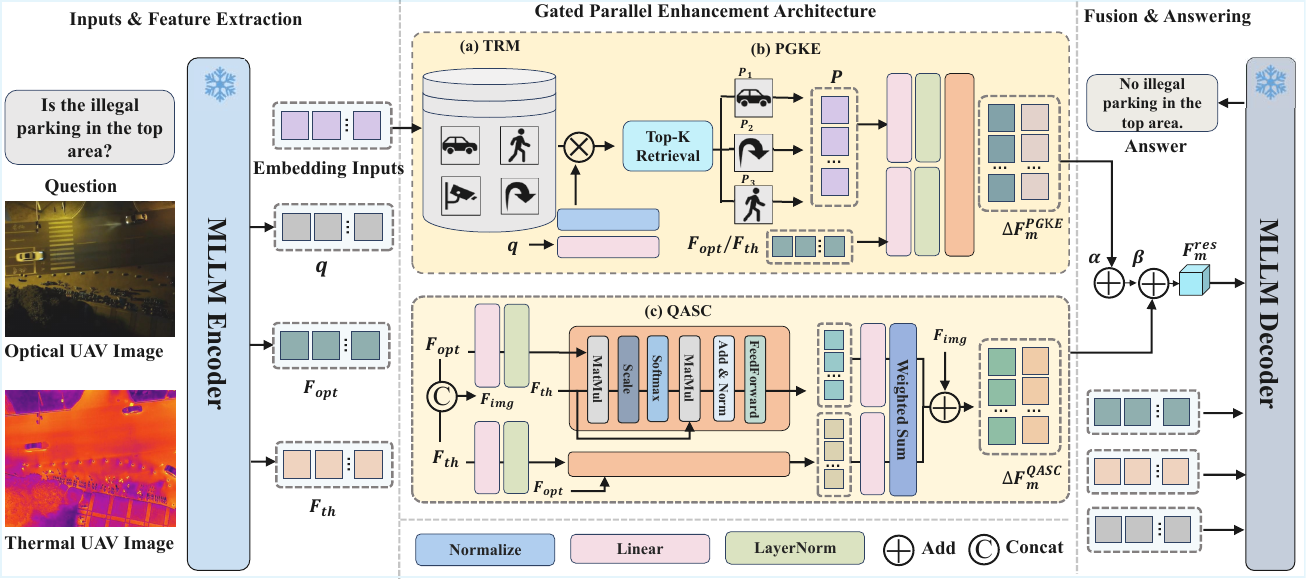}
\caption{Overall framework of MTCNet for multi-spectral UAV traffic VQA. The architecture adopts a Gated Parallel Residual paradigm in which the frozen, pre-trained MLLM visual features are adaptively augmented by domain-specific residual knowledge generated by the PGKE and QASC modules. The learnable gating parameters $\alpha$ and $\beta$ regulate the intensity of cognitive and multi-modal context injection.}
\label{fig:overall}
\end{figure*}

The overall architecture of MTCNet is illustrated in Fig.~\ref{fig:overall}. The framework consists of a frozen MLLM backbone coupled with two parallel, task-specific enhancement branches: the PGKE module and the QASC module.

Given a well-aligned pair of optical and TIR images, denoted as $I_{opt} \in \mathbb{R}^{H \times W \times 3}$ and $I_{th} \in \mathbb{R}^{H \times W \times 3}$, alongside a natural language query $Q$, MTCNet aims to generate a comprehensive textual answer $A$. To harness the generalized inferential capabilities of large-scale pre-training while adapting to the specificities of UAV traffic environments, we employ a frozen MLLM backbone (e.g., Qwen-VL) augmented with a non-invasive residual enhancement mechanism.

Let $\Phi_{\mathrm{enc}}(\cdot)$ denote the frozen visual encoder. Multi-scale visual features are first extracted from both modalities, then flattened and projected into a unified semantic embedding space, yielding the optical feature sequence $\mathbf{F}_{opt} \in \mathbb{R}^{L \times D}$ and the thermal feature sequence $\mathbf{F}_{th} \in \mathbb{R}^{L \times D}$, where $L$ is the sequence length of visual tokens and $D$ is the latent feature dimensionality. Concurrently, the query $Q$ is tokenized and embedded into a textual feature vector $\mathbf{q} \in \mathbb{R}^{D}$.

To prevent catastrophic forgetting of pre-trained knowledge, the backbone parameters are kept frozen while task-specific contextual information is injected via the parallel trainable branches. The refined visual representation $\mathbf{F}_{m}^{\mathrm{res}}$ for modality $m \in \{opt, th\}$ is formulated as:
\begin{equation}
\mathbf{F}_{m}^{\mathrm{res}} = \mathbf{F}_{m} + \alpha \cdot \Delta \mathbf{F}_{m}^{\mathrm{PGKE}} + \beta \cdot \Delta \mathbf{F}_{m}^{\mathrm{QASC}},
\label{eq:overall_fusion}
\end{equation}
where $\Delta \mathbf{F}_{m}^{\mathrm{PGKE}}$ and $\Delta \mathbf{F}_{m}^{\mathrm{QASC}}$ denote the residual feature tensors generated by the PGKE and QASC modules, respectively. The scalars $\alpha$ and $\beta$ are learnable gating parameters initialized at zero, enabling the network to autonomously regulate the injection intensity of cognitive prototypes and cross-modal context. The fused representations from both modalities are concatenated and fed into the frozen LLM decoder for autoregressive answer generation.

\subsection{Traffic Regulation Memory Construction}
\label{sec:cpb}

Existing MLLMs frequently generate imprecise descriptions when interpreting complex traffic violations, primarily due to the absence of domain-specific episodic memory. To bridge this gap, we establish an offline TRM, denoted as $\mathcal{M} \in \mathbb{R}^{N \times D_{p}}$, which stores $N$ high-level semantic prototypes derived from the training distribution. As illustrated in Fig.~\ref{fig:cpb_vis}, the construction of the TRM is carried out through three phases: semantic distillation, multi-modal visual grounding, and situation feature aggregation.

\subsubsection{Semantic Phrase Generation}
For each training triplet $(I_{opt}, I_{th}, Q, A)$, a text-only LLM is used to distill the core visual scenario into a concise, specific semantic phrase, designated as $P_{sem}$. Unlike unconstrained captions that tend to contain redundant information, the generation prompt explicitly instructs the model to identify critical traffic entities and their concurrent behaviors (e.g., "\textit{a white SUV executing an illegal U-turn across double yellow lines}"). This targeted distillation decouples the foundational visual-semantic elements from the grammatical structure of the raw QA pair, yielding a clean textual anchor for the subsequent localization phase.

\subsubsection{Multi-Modal Visual Grounding}
To spatially isolate the specified traffic event from the complex background, an open-set visual grounding framework (e.g., Qwen-VL or Grounding-DINO) is applied. Using the distilled $P_{sem}$ as the referential cue, the model predicts bounding boxes for both the optical ($B_{opt}$) and thermal ($B_{th}$) visual planes.

\begin{figure*}
\centering
\includegraphics[width=1.0\textwidth]{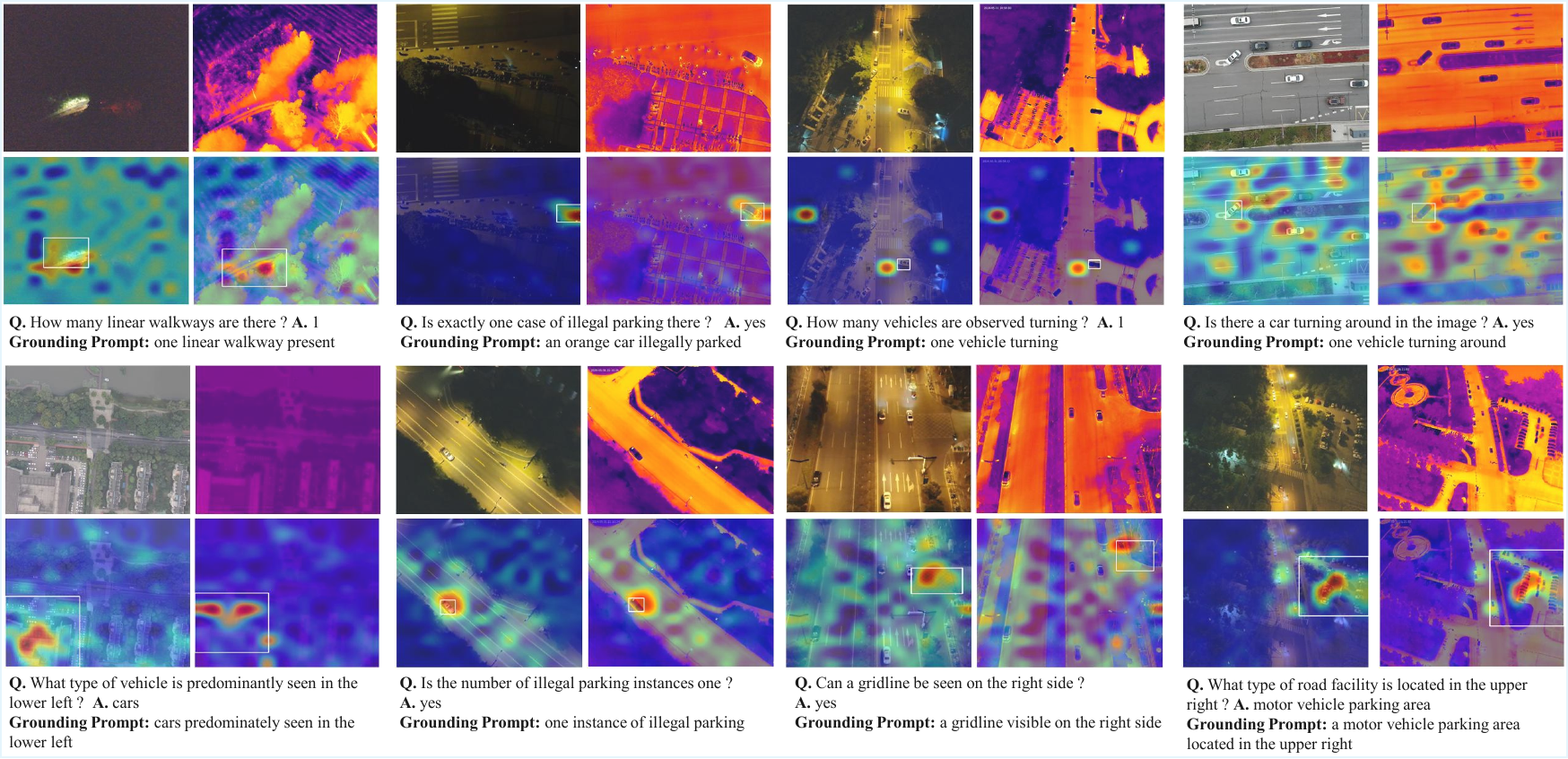}
\caption{Multi-modal visual grounding in the TRM construction pipeline. Grounding prompts generated from semantic phrase distillation are used to localize traffic entities and behaviors (e.g., linear walkways, vehicle turning). Red boxes indicate the extracted regions of interest, demonstrating accurate text-to-region alignment in both optical and thermal imagery.}
\label{fig:cpb_vis}
\end{figure*}

As shown in Fig.~\ref{fig:cpb_vis}, this process effectively translates abstract textual descriptors into precise, localized spatial coordinates. The visualizations confirm the robustness of our grounding strategy across diverse scenarios, successfully identifying static infrastructure such as "linear walkways" and "parking areas," as well as dynamic anomalies like "illegal parking" or "vehicle turning." Notably, the grounding prompts retain high efficacy even when the dominant visual features are defined primarily by thermal signatures. To further improve spatial robustness against single-modality occlusions (e.g., a vehicle clearly visible in TIR imagery but obscured in the optical view by shadows or foliage), we compute the union of the modality-specific bounding boxes to define a unified event Region of Interest (RoI), denoted as $B_{union}$:
\begin{equation}
B_{union} = B_{opt} \cup B_{th}.
\end{equation}

\subsubsection{Situation Feature Aggregation}
We propose an Epicenter Query mechanism to extract a representative prototype vector from within the localized $B_{union}$ region. Let $\mathbf{F}_{concat} = [\mathbf{F}_{opt}; \mathbf{F}_{th}] \in \mathbb{R}^{L \times 2D}$ denote the concatenated multi-modal feature map. The epicenter query $\mathbf{q}_{epi}$ is derived by mean-pooling over the features within the spatial center neighborhood of $B_{union}$. The cognitive prototype $\mathbf{s} \in \mathbb{R}^{2D}$ is then computed via a weighted aggregation:
\begin{equation}
\mathbf{s} = \sum_{i=1}^{L} \frac{\exp(\mathbf{F}_{concat}^{(i)} \cdot \mathbf{q}_{epi}^\top / \tau)}{\sum_{j=1}^{L} \exp(\mathbf{F}_{concat}^{(j)} \cdot \mathbf{q}_{epi}^\top / \tau)} \mathbf{F}_{concat}^{(i)},
\end{equation}
where $\tau$ is a temperature scaling factor. This aggregation yields a refined prototype $\mathbf{s}$ that encapsulates the core visual-semantic essence of the localized traffic situation. The collection of these vectors forms the TRM $\mathcal{M}$, which serves as an external knowledge base for subsequent complex cognitive tasks.

\subsection{Prototype-Guided Knowledge Embedding}
\label{sec:pgke}

The PGKE module bridges the gap between the current visual input and the external domain knowledge stored in the TRM. It operates via a retrieve-and-align strategy designed to generate the cognitive residual feature $\Delta \mathbf{F}_{m}^{\mathrm{PGKE}}$.

\textbf{Prototype Retrieval.}
The current textual question embedding $\mathbf{q}$ is first linearly projected into the prototype space using a transformation matrix $\mathbf{W}_q$. The system computes the cosine similarity between this projected question vector and all prototypes in $\mathcal{M}$, retrieving the top-$K$ most semantically relevant prototypes to form a support set $\mathcal{P}_{ret} \in \mathbb{R}^{K \times D}$:
\begin{equation}
\mathcal{P}_{ret} = \mathrm{TopK}\left( \frac{(\mathbf{q}\mathbf{W}_q) \mathcal{M}^\top}{\|\mathbf{q}\mathbf{W}_q\| \|\mathcal{M}\|} \right).
\end{equation}
These retrieved prototypes serve as reference anchors, supplying critical feature signatures for recognizing abstract traffic anomalies (e.g., recalling the visual pattern associated with a "rear-end collision").

\begin{figure}
\centering
\includegraphics[width=0.45\textwidth]{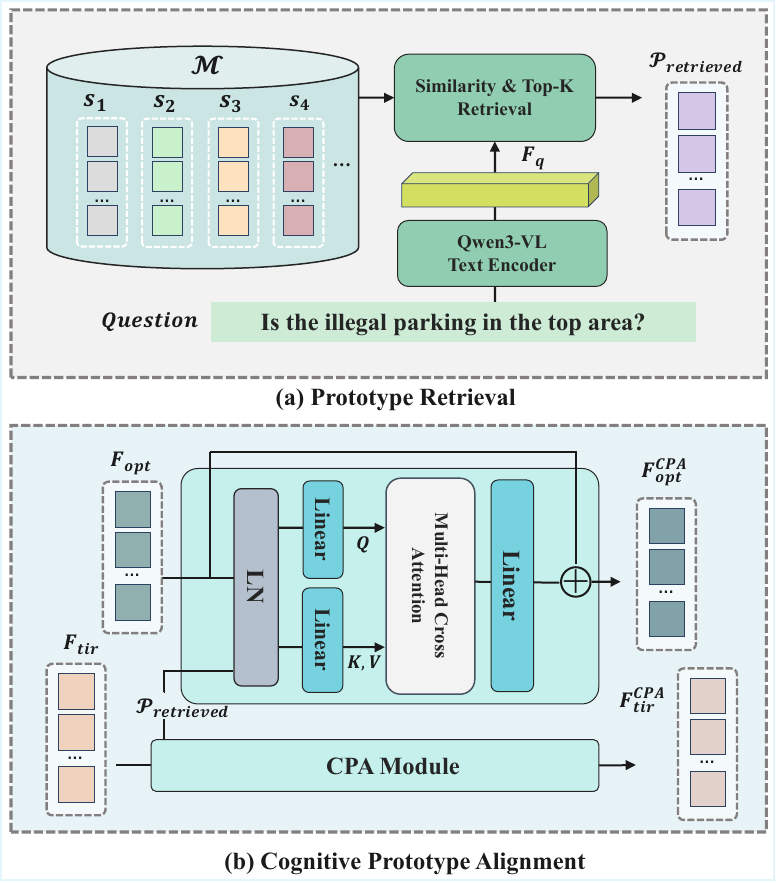}
\caption{Internal architecture of the PGKE module. The module performs question-guided similarity retrieval to identify the top-$K$ most relevant prototypes from the TRM. These prototypes serve as keys and values in a Multi-Head Cross-Attention mechanism, injecting situational domain knowledge into the visual feature streams as an optimized residual increment $\Delta \mathbf{F}^{\mathrm{PGKE}}$.}
\label{fig:cpa}
\end{figure}

\textbf{Cognitive Prototype Alignment.}
The detailed mechanism of the PGKE module, including the retrieve-and-align phase via multi-head cross-attention, is illustrated in Fig.~\ref{fig:cpa}. We employ a Multi-Head Cross-Attention framework to inject the retrieved regulatory knowledge into the visual feature hierarchy. For modality $m$, the visual features $\mathbf{F}_m$ serve as the Query ($\mathbf{Q}_{pgke}$), while the retrieved prototypes $\mathcal{P}_{ret}$ serve as both the Key ($\mathbf{K}_{pgke}$) and Value ($\mathbf{V}_{pgke}$):
\begin{align}
\mathbf{Q}_{pgke} &= \mathrm{LN}(\mathbf{F}_m)\mathbf{W}_Q^{pgke}, \\
\mathbf{K}_{pgke}, \mathbf{V}_{pgke} &= \mathrm{LN}(\mathcal{P}_{ret})\mathbf{W}_K^{pgke}, \mathrm{LN}(\mathcal{P}_{ret})\mathbf{W}_V^{pgke}.
\end{align}
The cognitive residual is then computed as:
\begin{equation}
\Delta \mathbf{F}_{m}^{\mathrm{PGKE}} = \mathrm{softmax}\left(\frac{\mathbf{Q}_{pgke}\mathbf{K}_{pgke}^\top}{\sqrt{d_k}}\right)\mathbf{V}_{pgke}\mathbf{W}_O^{pgke}.
\end{equation}
Through this mechanism, the visual features are aligned with high-level domain knowledge, substantially enhancing the network's discriminative ability in fine-grained cognitive evaluations.

\subsection{Quality-Aware Spectral Compensation}
\label{sec:qasc}

To overcome the limitations of basic feature concatenation, we design the QASC module (depicted in Fig.~\ref{fig:overall} (c)), which facilitates bidirectional, dynamically balanced context exchange between optical and thermal features.

In contrast to simple concatenation or element-wise addition, QASC leverages a symmetric bidirectional attention mechanism to exchange complementary context between modalities. For the fusion direction from thermal to optical (compensating $\mathbf{F}_{opt}$ using information from $\mathbf{F}_{th}$), the query is derived from the optical features, while the key and value pair are derived from the thermal features:
\begin{equation}
\Delta \mathbf{F}_{opt}^{\mathrm{QASC}} = \mathrm{MHCA}(\mathbf{Q}=\mathbf{F}_{opt}, \mathbf{K}=\mathbf{F}_{th}, \mathbf{V}=\mathbf{F}_{th}).
\end{equation}
Conversely, the thermal features are simultaneously enriched by the optical context:
\begin{equation}
\Delta \mathbf{F}_{th}^{\mathrm{QASC}} = \mathrm{MHCA}(\mathbf{Q}=\mathbf{F}_{th}, \mathbf{K}=\mathbf{F}_{opt}, \mathbf{V}=\mathbf{F}_{opt}).
\end{equation}
Standard Layer Normalization (LN) and Feed-Forward Networks (FFNs) are applied following the standard Transformer blocks. This symmetric architecture ensures that under adverse conditions (e.g., deep darkness or dense fog), the modality with superior signal quality (predominantly thermal) effectively guides the refinement of the degraded modality (optical), thereby maximizing the resilience and fidelity of the final joint representation.

\subsection{Loss Function}
Following standard MLLM training protocols, MTCNet is optimized via an auto-regressive language modeling objective. Given the sequence of ground-truth answer tokens $\mathbf{y} = \{y_1, y_2, ..., y_T\}$, the training objective is to minimize the negative log-likelihood:
\begin{equation}
\mathcal{L} = - \sum_{t=1}^{T} \log P(y_t | y_{<t}, \mathbf{F}_{opt}^{\mathrm{res}}, \mathbf{F}_{th}^{\mathrm{res}}, \mathbf{q}; \Theta),
\end{equation}
where $\Theta$ encompasses all trainable parameters within the PGKE and QASC modules, including the gating scalars. The parameters of the pre-trained visual encoder and LLM backbone remain frozen throughout optimization.

\section{Traffic-VQA Dataset}
\label{sec:dataset}

\begin{figure*}
\centering
\includegraphics[width=1.0\textwidth]{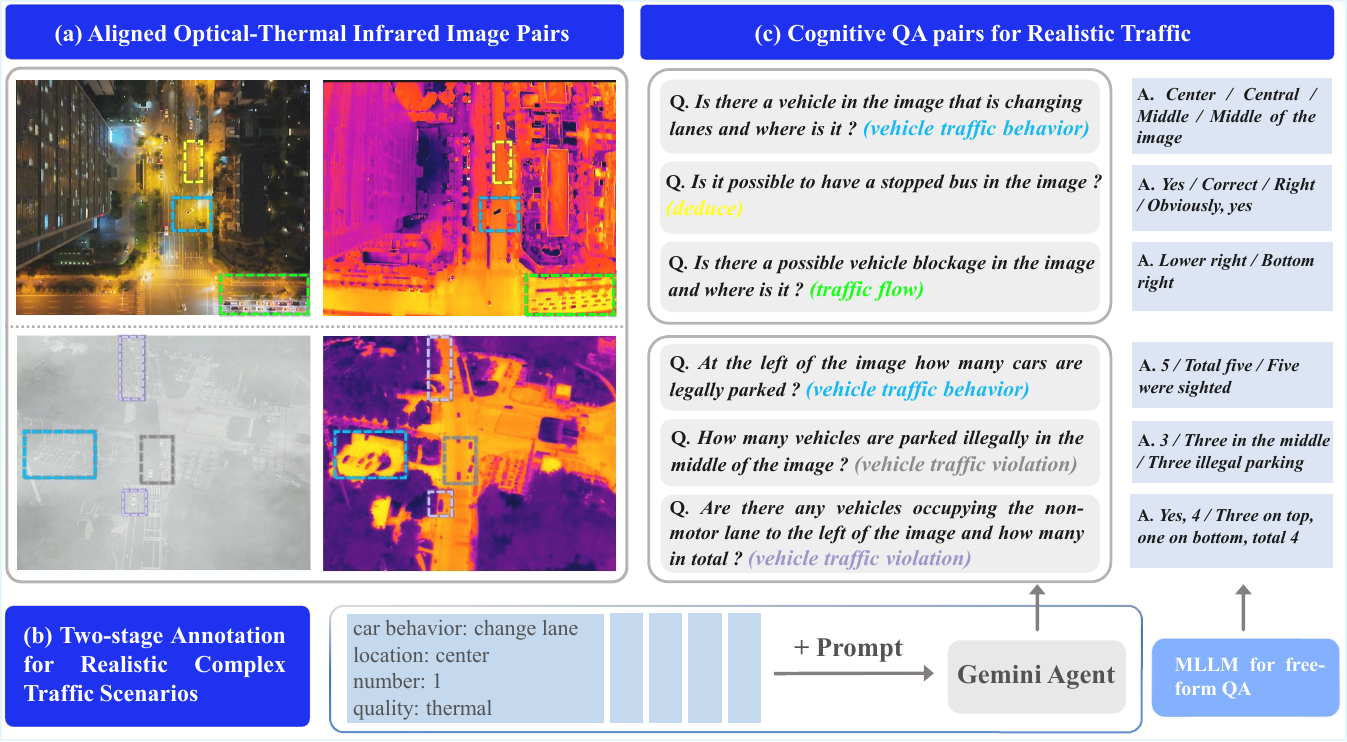}
\caption{Illustrative examples from the Traffic-VQA dataset. (a) Synchronized and co-registered optical and TIR UAV image pairs across diverse urban traffic settings. (b) Examples of challenging cognitive question-answer pairs that require deep situational understanding, such as identifying traffic violations and inferring latent behavioral risks.}
\label{fig:dataset_samples}
\end{figure*}

\begin{figure*}
  \centering
  \includegraphics[width=0.98\textwidth]{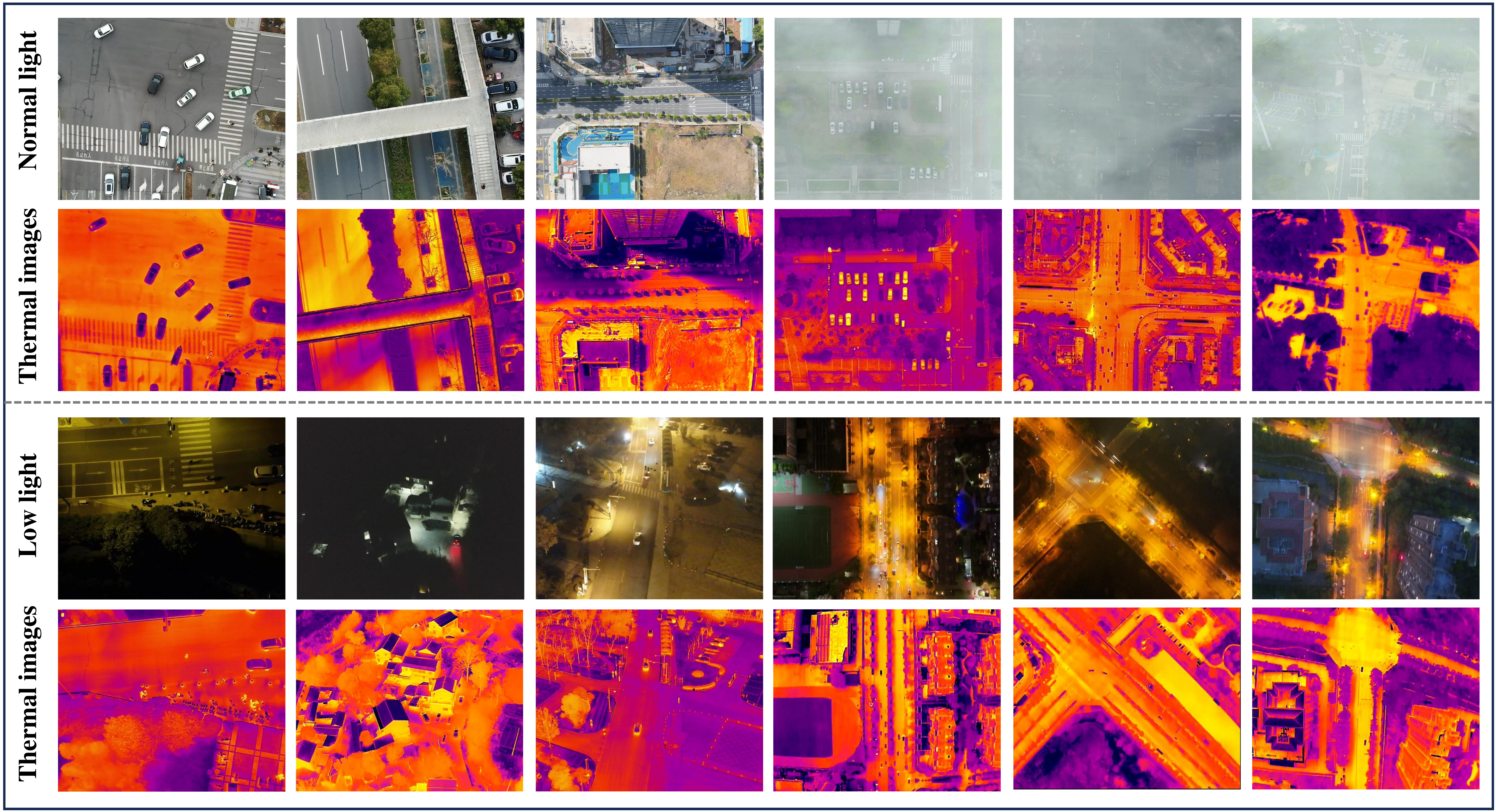}
  \caption{Sample aligned image pairs from the proposed Traffic-VQA dataset. The dataset spans diverse illumination conditions (e.g., daylight, low-light, and nighttime) alongside adverse weather scenarios (e.g., fog), providing well-aligned OPT and TIR images to support research in all-weather traffic understanding.}
  \label{fig:dataset_samples2}
\end{figure*}

To bridge the gap between elementary perception and complex cognitive interpretation in traffic surveillance, we construct Traffic-VQA, the first large-scale OPT-TIR benchmark dedicated to cognitive traffic understanding. Unlike existing datasets that predominantly focus on rudimentary object detection or counting in well-lit environments, Traffic-VQA is designed to challenge models with intricate cognitive tasks under diverse and adverse environmental conditions.

\subsection{Review of Existing Datasets}
\label{ssec:existing_datasets}
While RSVQA has evolved considerably, existing benchmarks exhibit clear limitations when applied to the cognitive interpretation of complex traffic environments. We categorize the existing dataset landscape into three distinct groups:

\myparagraph{Foundational Perception Benchmarks.}
Early datasets established the groundwork for fundamental remote sensing perception. The seminal RSVQA \cite{RSVQA} and RSVQAxBEN \cite{RSVQA_Meets} predominantly employed template-based questions programmatically generated from OpenStreetMap (OSM) data overlaying low-resolution Sentinel-2 or aerial imagery. Although large in scale, the generated queries are structurally rigid (e.g., "Is there a building?") and lack semantic diversity. RSIVQA \cite{rsivqa} and the Open-Ended dataset \cite{VQA-TextRS} introduced human-annotated questions to increase natural language variety, yet they remain fundamentally constrained to static object recognition within optimal optical imagery.

\myparagraph{Task-Specific and Disaster Response Benchmarks.}
A second group focuses on specific high-stakes operational scenarios. FloodNet \cite{floodnet} and SAM-VQA \cite{SAM-VQA} exclusively address post-disaster damage assessment, requiring models to evaluate flood impacts or structural building damage. Similarly, CDVQA \cite{CDVQA} introduces VQA specific to change detection on multitemporal images. While these datasets, alongside methods like TGFNet \cite{text-guided}, advance domain-specific understanding, their exclusive reliance on RGB sensors renders them unsuitable for continuous traffic monitoring, where nighttime and adverse weather capabilities are essential.

\myparagraph{Cognition-Oriented and Large-Scale Benchmarks.}
Recent efforts have aimed to align datasets with the capabilities of MLLMs. CRSVQA \cite{MQVQA} introduces complex, multi-tiered questions to benchmark advanced question-driven systems. RSGPT \cite{rsgpt} establishes a benchmark for simultaneous image captioning and VQA. EarthVQA \cite{earthvqa} emphasizes complex relational analysis for urban planning applications, whereas LRS-VQA \cite{When_Large} targets gigapixel-level ultra-large imagery interpretation. EarthGPT \cite{earthgpt} integrates a diverse set of tasks but still relies fundamentally on optical data. Critically, none of these benchmarks provide well-aligned multi-modal (optical and TIR) data for robust all-weather perception, nor do they target the fine-grained cognitive tasks (e.g., traffic violation detection, vehicle behavior analysis) that are central to intelligent transportation systems.

In contrast, Traffic-VQA addresses these limitations by introducing the aligned TIR modality alongside a hierarchical cognitive structure tailored for dynamic, real-time traffic cognition. Table~\ref{tab:comparison} presents a comprehensive comparison between Traffic-VQA and existing datasets.

\begin{table*}[t]
    \centering
    \caption{Comparison of Traffic-VQA with leading VQA datasets utilizing overhead imagery. (OPT: Optical, SAR: Synthetic Aperture Radar, TIR: Thermal Infrared)}
    \label{tab:comparison}
    \begin{adjustbox}{width=\textwidth, keepaspectratio}
    \begin{tabular}{l | c c c c c c l}
        \toprule
        \textbf{Dataset} & \header{Num \\ Img} & \header{Num \\ QA} & \header{Num Qst \\ Type} & \header{Main QA \\ Generation} & \header{Incl. \\ Cognitive} & \header{Modal} & \header{Key Tasks} \\
        \midrule
        RSVQA-LR~\cite{RSVQA}       & 772    & 77k        & 5 & Template          & \XSolidBrush    & OPT & Counting, Presence \\
        RSVQA-HR~\cite{RSVQA}       & 10,659  & 1,066k     & 5 & Template          & \XSolidBrush    & OPT & Counting, Presence \\
        VQA-TextRS~\cite{VQA-TextRS} & 2,144   & 6.2k       & 4 & Manual            & \XSolidBrush    & OPT & Object Recognition \\
        RSIVQA~\cite{rsivqa}        & 37,264  & 111k       & 9 & Template          & \XSolidBrush    & OPT & Object Recognition \\
        CRSVQA~\cite{MQVQA}         & 4,639   & 4.6k       & 3 & Manual            & \XSolidBrush    & OPT & Multistep Analysis \\
        FloodNet-VQA~\cite{floodnet} & 2,188   & 7.4k       & 4 & Manual            & \XSolidBrush    & OPT & Disaster Assessment \\
        SAM-VQA~\cite{SAM-VQA}      & 2,348   & 10.5k      & 7 & Template          & \XSolidBrush    & OPT & Damage Analysis \\
        CDVQA~\cite{CDVQA}          & 4,662   & $>$122k    & 5 & Template          & \CheckmarkBold  & OPT & Change Detection \\
        RSIEval~\cite{rsgpt}        & 100     & 0.9k       & 4 & Manual            & \CheckmarkBold  & OPT & Captioning \& VQA \\
        EarthVQA~\cite{earthvqa}    & 6,000   & 209k       & 3 & Manual + Template & \CheckmarkBold  & OPT & Relational Analysis \\
        OSVQA~\cite{text-guided}         & 6,008 pairs & 1,037k & 16 & Manual + Template & \CheckmarkBold  & OPT \& SAR & Multi-modal Perception \\
        \midrule
        \textbf{Traffic-VQA (Ours)} & \textbf{8,180 pairs} & \textbf{1,301k} & \textbf{31} & \textbf{Manual + LLM} & \textbf{\CheckmarkBold} & \textbf{OPT \& TIR} & \textbf{Traffic Cognition} \\
        \bottomrule
    \end{tabular}
    \end{adjustbox}
\end{table*}

\begin{figure*}
\centering
\includegraphics[width=1.0\textwidth]{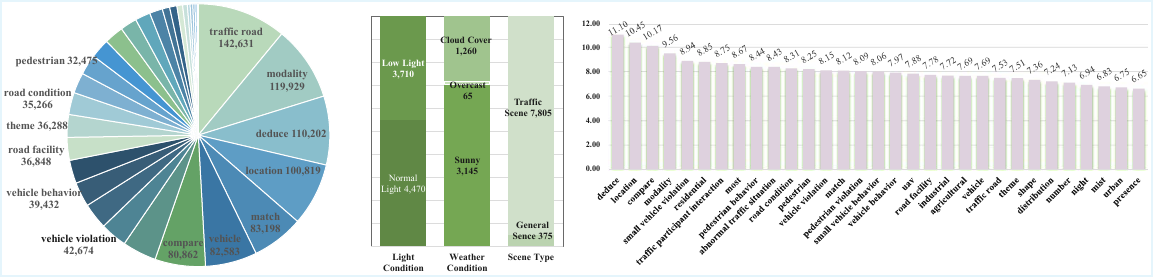}
\caption{Statistical distribution of the Traffic-VQA dataset. Left: Breakdown of the 8,180 image pairs by illumination conditions, weather, and scene type. Right: Distribution of the 31 question types, highlighting the significant proportion of high-level cognitive tasks (20.7\%) and specialized multi-modal queries.}
\label{fig:stats}
\end{figure*}

\begin{figure}
    \centering
    \includegraphics[width=1.0\columnwidth]{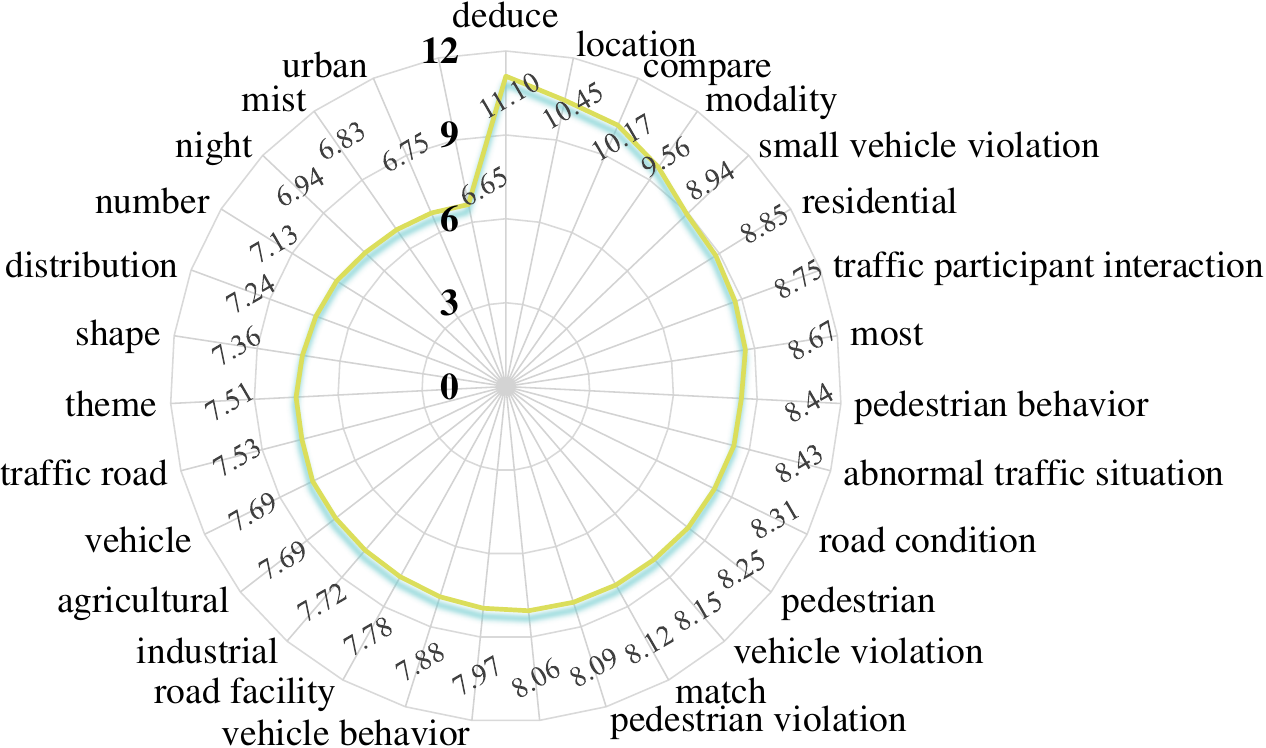}
    \caption{Distribution of average question length (in words) across varying question types within the Traffic-VQA dataset.}
    \label{fig:question_length}
\end{figure}

\subsection{Data Construction Pipeline}
\label{ssec:data_construction}
Constructing a large-scale, cognitively rich, and robustly multi-modal benchmark requires rigorous protocols to ensure both the visual fidelity of sensor data and the semantic integrity of linguistic annotations. To this end, we designed a systematic Human-in-the-Loop LLM-Empowered Construction Pipeline that minimizes manual annotation effort while expanding the structural diversity and cognitive complexity of the resulting queries. The construction workflow proceeds through three stages: Stage 1: Hardware-Synchronized Data Acquisition, Stage 2: Structured Attribute Annotation by Experts, and Stage 3: LLM-Empowered QA Generation.

\myparagraph{Stage 1: Hardware-Synchronized Data Acquisition.}
To ensure spatial and temporal consistency, we deployed the DJI M300 RTK drone platform equipped with the Zenmuse H20T integrated payload. This payload houses a calibrated wide-angle optical sensor alongside a radiometric thermal camera, enabling hardware-synchronized capture of optical and TIR imagery. This integrated hardware approach eliminates the temporal desynchronization and spatial misalignment artifacts common in conventional multi-sensor arrays.
Data collection was carried out across diverse urban environments, including high-density arterial roads, complex signalized intersections, and highway on/off ramps. To ensure all-weather perceptual robustness, flights were conducted under a wide range of illumination conditions (daylight, dusk/dawn, and nighttime) as well as adverse weather (dense fog).
Post-acquisition, a manual screening protocol was enforced. Image pairs exhibiting motion blur, excessive occlusion, or corrupted sensor data were discarded, yielding a curated dataset of 8,180 high-quality, well-aligned image pairs.

\myparagraph{Stage 2: Structured Attribute Annotation by Experts.}
Relying solely on direct manual annotation for large-scale QA datasets frequently results in simplistic and repetitive linguistic constructs. To address this, we adopted an attribute-centric annotation paradigm.
We first defined a comprehensive Traffic Cognition Ontology covering granular object categorizations (e.g., sedan, heavy truck, pedestrian), dynamic behavioral classifications (e.g., executing a turn, exceeding speed limits, queuing), and environmental state attributes (e.g., visibility metrics, road surface conditions).
Certified domain experts were then tasked with annotating the imagery according to this ontology. Rather than drafting free-form questions, the experts focused on extracting structured, ground-truth metadata:
\begin{itemize}
    \item \textbf{Object-Level.} Precise bounding box annotations paired with attribute tags for individual traffic participants.
    \item \textbf{Scene-Level.} Global tags classifying weather conditions, ambient lighting, road type, and traffic density.
    \item \textbf{Event-Level.} Identification of specific vehicular violations or behavioral anomalies (e.g., "vehicle illegally positioned on pedestrian sidewalk").
\end{itemize}
This structured metadata repository serves as the factual foundation for the subsequent generative phase.

\myparagraph{Stage 3: LLM-Empowered QA Generation and Verification.}
Leveraging the capabilities of LLMs, we automatically synthesized the structured expert annotations into complex, natural language QA pairs. This generation phase operates through two parallel processes:
\begin{enumerate}
    \item \textbf{Programmatic Generation for Precision.} For objective queries requiring exact numerical counting or binary existence verification (e.g., "How many standard passenger cars are currently visible?"), we applied rule-based templates populated directly from the verified annotation database, ensuring accuracy.
    \item \textbf{LLM-Based Cognitive Context Expansion.} For advanced cognitive tasks (e.g., "Is the current traffic configuration indicative of a dangerous anomaly?"), the structured attributes were fed into GPT-4 using specialized, constrained prompts. These prompts guided the LLM to formulate multi-hop comprehension questions, logically connecting cause-and-effect relationships and demanding comparative analysis of visual features across both optical and TIR modalities.
\end{enumerate}
Finally, a Human-in-the-Loop Verification mechanism was applied. An independent team of verification experts sampled the generated QA pairs, correcting minor logical inconsistencies and filtering out ambiguous or poorly structured samples. This pipeline ultimately produced over 1.3 million high-quality QA pairs with diverse linguistic structures, far exceeding the scale and cognitive depth achievable through purely manual annotation.

\begin{figure*}
  \centering
  \includegraphics[width=0.98\textwidth]{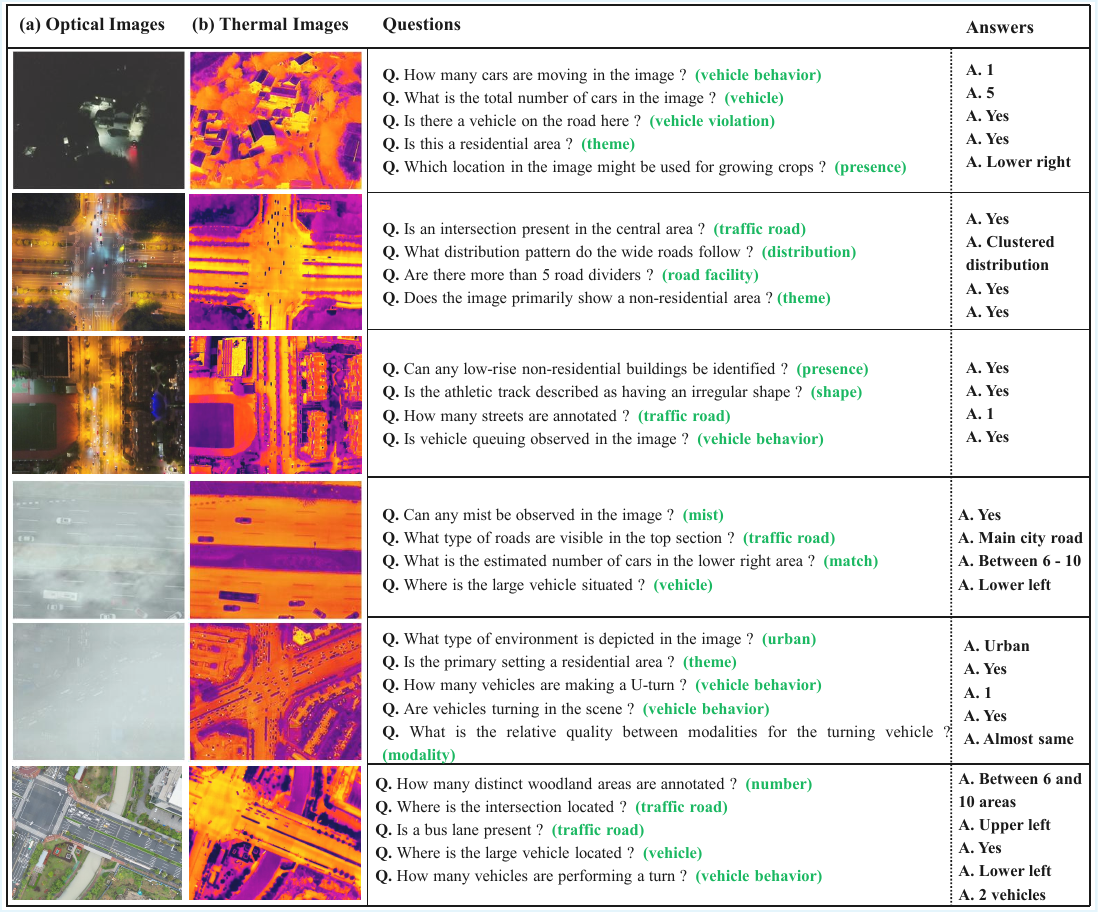}
  \caption{Representative question-answer samples from Traffic-VQA. The dataset covers diverse query types beyond object recognition, requiring reasoning over spatial relations, vehicle behaviors, traffic-rule compliance, and scene context (e.g., queuing, environment type).}
  \label{fig:dataset_qa_samples}
\end{figure*}

\begin{table*}[!t]
    \centering
    \vspace{-2mm}

    \begin{minipage}[t]{0.67\textwidth}
        \centering
        \captionsetup{width=\linewidth}
        \small
        \caption{Categorization of the 31 question types in the Traffic-VQA dataset, divided into cognitive, perceptual, and multi-modal domains.}
        \label{question_type}
        \begin{adjustbox}{width=\linewidth, totalheight=\textheight, keepaspectratio}
        \begin{tabular}{ll p{8cm}}
            \toprule
            \textbf{Question Type} & \textbf{Object Type}  & \textbf{Specific Question Type} \\
            \midrule
            \multirow{4}{*}{Cognitive} & \multirow{4}{*}{Traffic Participant}  & Abnormal Traffic Situation, Vehicle Violations, Small Vehicle Violation, Pedestrian Violation, Vehicle Behavior, Small Vehicle Behavior, Deduce, Pedestrian Behavior, Traffic Participant Interaction, Road Condition \\
            \midrule
            \multirow{4}{*}{Perceptual} & Traffic Participant & Traffic Road, Vehicle, Pedestrian, Road Facility \\
                                                \cmidrule(lr){2-3}
                                                & \multirow{3}{*}{Total Object} & Compare, Presence, Location, Number, Shape, Most, Distribution, Residential, Agricultural, Industrial, Mist, Night, Urban, Theme, UAV \\
            \midrule
            Modal                               & Total Object                    & Match, Modality \\
            \bottomrule
        \end{tabular}
        \end{adjustbox}
    \end{minipage}
    \hfill
    \begin{minipage}[t]{0.29\textwidth}
        \centering
        \captionsetup{width=\linewidth}
        \small
        \caption{Statistical overview of the Traffic-VQA dataset.}
        \label{tab:dataset_statistics}
        \resizebox{\linewidth}{!}{
        \setlength{\tabcolsep}{2pt}
        \renewcommand{\arraystretch}{0.2}
        \begin{tabular}{l r}
            \toprule
            \textbf{Statistics} & \textbf{Numbers} \\
            \midrule
            Total QAs & 1,301,466 \\
            Total OPT-TIR pairs & 8,180 \\
            Avg QA on img pair & 159.1 \\
            \midrule
            Max question length & 28 \\
            Avg question length & 8.5 \\
            Question vocab size & 24,357 \\
            Total question types & 31 \\
            \midrule
            Max answer length & 16 \\
            Avg answer length & 1.2 \\
            Answer vocab size & 4,418 \\
            Total unique answers & 7,644 \\
            \bottomrule
        \end{tabular}
        }
    \end{minipage}
    \vspace{-1mm}
\end{table*}

\subsection{Dataset Analysis}
\label{ssec:dataset_analysis}
The Traffic-VQA dataset is characterized by its large scale, diversity, and cognitive depth. We present a detailed statistical analysis of the distribution of question types, environmental conditions, and linguistic complexity.

\myparagraph{Statistical Distribution of Question Types.}
The finalized dataset comprises 1,301,466 verified QA pairs derived from 8,180 image pairs. The queries are distributed across distinct functional categories. The most frequent types include Traffic Road (142,631 queries) and Modality (119,929 queries), reflecting the dataset's focus on structural road understanding and cross-modal perception.
Crucially, the dataset emphasizes higher-order cognition: Deduce alone includes 110,202 samples, and Location comprehension contributes 100,819 samples. Two other high-impact categories are Compare Vehicle (approximately 80k) and Vehicle Violation (42,674), both directly relevant to automated traffic surveillance. This distribution pushes evaluation beyond basic object presence toward situation understanding and rule-aware interpretation.

\myparagraph{Environmental Diversity.}
To evaluate algorithmic robustness, Traffic-VQA incorporates a wide range of challenging environmental conditions. The dataset includes 4,470 image pairs captured under normal light, contrasted with 3,710 pairs acquired under low light, alongside dedicated subsets for Sunny (3,145) and Cloud Cover (1,260) scenarios.
This structured variety prevents models from overfitting to idealized conditions, instead requiring the learning of invariant feature representations across radical illumination shifts (e.g., transitioning from optical-dominant features during daylight to TIR-dominant features at night).

\myparagraph{Linguistic Complexity and Length.}
We quantified linguistic complexity by computing the average question length across different question types. The distribution shows a natural variance, with average lengths ranging from 6 to 12 words.
As expected, cognition-heavy queries tend to be longer; for example, questions related to Mist exhibit the highest average length (11.10 words), followed by Compare tasks (10.45 words) and Deduce queries (7.36 words).
Simpler perception tasks, such as Urban classification, average approximately 6.76 words. This correlation between task difficulty and question length confirms that Traffic-VQA challenges MLLMs to process complex, natural language queries that closely reflect the nuanced interactions of real-world operators.

\subsection{Dataset Challenges}
\label{ssec:challenges}
The Traffic-VQA dataset introduces several key challenges to contemporary VQA architectures and MLLM methodologies, highlighting important directions for future research.

\begin{enumerate}[label=\arabic*), leftmargin=*]
    \item \textbf{Cross-Modality Semantic Alignment.}
    Effectively fusing information from optical and TIR modalities remains a significant challenge. TIR images lack color information but provide strong spatial contrast for heat-emitting objects, while optical images offer rich textural detail but fail under low-light conditions. Models must learn to dynamically weight the relevance of each modality depending on the context (e.g., prioritizing TIR data at night) and align inherently inconsistent semantic features across these disparate sensors.

    \item \textbf{Fine-Grained Object Cognition.}
    UAV imagery typically captures a large field of view, making the constituent objects very small relative to the overall image dimensions. Performing deep cognitive analysis on the behavior of a single small vehicle (e.g., determining "Is the third car in the far-left lane actively executing a turn?") requires precise attention mechanisms and fine-grained spatial localization capabilities, which are difficult to achieve with standard vision encoders pre-trained on ground-level imagery.

    \item \textbf{Robustness to Environmental Degradation.}
    The dataset includes a substantial proportion of foggy and low-light scenarios where standard visual features are severely degraded. Traditional VQA models typically suffer significant performance drops when confronted with such extreme domain shifts. Traffic-VQA therefore serves as a benchmark for developing robust representation learning methods that maintain high cognitive performance regardless of adverse weather or lighting conditions.
\end{enumerate}

In summary, Traffic-VQA provides a comprehensive platform for advancing multi-modal cognitive intelligence in the domain of intelligent transportation systems.

\section{Experiments}
\label{sec:experiments}

In this section, we present an empirical evaluation of the proposed MTCNet framework on the Traffic-VQA benchmark. We first describe the experimental configuration and implementation details (Section~\ref{sec:implementation}). We then provide a quantitative comparison against state-of-the-art (SOTA) MLLMs (Section~\ref{sec:quantitative}), focusing on performance gains attributable to our cognitive and perceptual modules. Finally, ablation studies (Section~\ref{sec:ablation}) verify the individual contributions of the PGKE and QASC modules, complemented by a fine-grained analysis across question types.

\subsection{Implementation Details}
\label{sec:implementation}

\myparagraph{Model Configuration.}
We employ Qwen3-VL-8B as the frozen backbone, selected for its strong visual-linguistic alignment capabilities. The input resolution for both optical and TIR images is set to $640 \times 512$ pixels. To preserve the generalization capabilities of the pre-trained MLLM and ensure parameter efficiency, the core vision encoder and LLM backbone remain entirely frozen. Only the parameters of the Gated Parallel Residual Architecture---comprising the PGKE and QASC modules---are updated during training.

\myparagraph{Training Protocols.}
All models are implemented in PyTorch and trained on NVIDIA RTX 4090 GPUs. We use the AdamW optimizer with a cosine learning rate scheduler (initial learning rate $1 \times 10^{-4}$) and a batch size of 16. To evaluate the efficacy and adaptability of our architecture, we adopt two training protocols:
\begin{enumerate*}[label=(\roman*)]
    \item \textbf{Few-Shot Learning.} To demonstrate the effectiveness of prototype-guided knowledge injection under data-scarce conditions, the model is fine-tuned on a randomly sampled subset of 10,000 examples.
    \item \textbf{Full Fine-Tuning.} To establish the upper performance bound of the architecture, the model is trained on the complete Traffic-VQA training set.
\end{enumerate*}

\myparagraph{Evaluation Metrics.}
Following standard VQA evaluation protocols, we report both Accuracy (Acc) and CIDEr (C) metrics. The evaluation structure distinguishes between \textit{Cognitive} and \textit{Perceptual} tasks, and \textit{multi-modal} versus \textit{single-modal} inputs, to highlight the cognitive depth and environmental robustness introduced by the proposed method.

\begin{table*}[t]
\centering
\caption{Quantitative comparison on the Traffic-VQA test set. All numbers are \textbf{Acc} (\%). \textbf{Bold} indicates the best result and \underline{underline} indicates the second-best result \emph{within the same modality} for each metric.}
\label{tab:traffic_vqa_full_acc}
\renewcommand{\arraystretch}{1.15}
\setlength{\tabcolsep}{4pt}
\footnotesize
\begin{tabular}{l l|c c c|c c c|c c c}
\toprule
\textbf{Model} & \textbf{Source} & \textbf{OA} & \header{\textbf{Traffic} \\ \textbf{Cognition}} & \header{\textbf{Traffic} \\ \textbf{Perception}} & \header{\textbf{Veh.} \\ \textbf{Violation}} & \header{\textbf{Veh.} \\ \textbf{Behavior}} & \textbf{Veh.} & \header{\textbf{Ped.} \\ \textbf{Violation}} & \header{\textbf{Ped.} \\ \textbf{Behavior}} & \textbf{Ped.} \\
\midrule
\multicolumn{11}{c}{\textit{\color{gray} OPT modality}} \\ \hline
MiniCPM-V~\cite{MiniCPM-V} & arXiv 2024 & 45.27 & 39.7 & 45.43 & 37.12 & 39.57 & 47.57 & 40.13 & 44.82 & 41.22 \\
MiniGPT-v2~\cite{minigpt-v2} & arXiv 2023 & 36.04 & 35.3 & 35.95 & 33.74 & 37.38 & 37.96 & 29.54 & 39.05 & 40.14 \\
DeepSeek-VL~\cite{deepseek-vl} & arXiv 2024 & 42.64 & 38.62 & 41.51 & 37.01 & 37.79 & 44.38 & 41.23 & 42.92 & 34.86 \\
GeoPix~\cite{geopix} & GRSM 2025 & 43.59 & 38.13 & 40.48 & 37.16 & 39.49 & 45.1 & 38.07 & 43.94 & 41.86 \\
GeoChat~\cite{geochat} & CVPR 2024 & 44 & 41.24 & 44.39 & 41.98 & 41.38 & 45.32 & 42.97 & 42.56 & 46.05 \\
Falcon~\cite{falcon} & arXiv 2025 & 34.22 & 38.27 & 40.41 & 37.57 & 32.45 & 37.83 & 43.29 & 46.42 & 37.84 \\
Qwen2.5-VL-7B~\cite{qwen2.5-vl} & Ali 2025 & 43.69 & 40.04 & 60.61 & 39.5 & 39.02 & 47.19 & 48.97 & 44.53 & 43.4 \\
Qwen3-VL-8B (Base)~\cite{qwen3-vl} & Ali 2025 & 47.56 & 35.11 & 46.79 & 34.76 & 32.67 & 56.01 & 31.45 & 40.02 & 42.95 \\
GPT-4o~\cite{gpt4o} & OpenAI 2024 & 64.81 & 71.92 & 72.33 & 74.15 & 71.22 & 73.27 & 64.3 & 73.8 & 62.46 \\
Gemini-2.5-Flash~\cite{gemini2.5pro} & Google 2025 & 59.51 & 66.7 & 69.77 & 65 & 66.8 & 73.34 & 63.35 & 70.22 & 62.17 \\
Gemini-2.5-Pro~\cite{gemini2.5pro} & Google 2025 & 56.17 & 65.98 & 63.18 & 64.95 & 64.16 & 67.44 & 56.24 & 62.85 & 58.27 \\
MTCNet (Few-Shot) & - & 47.56 & 46.79 & 35.11 & 34.76 & 32.67 & 56.01 & 42.95 & 40.02 & 31.45 \\
\midrule
\multicolumn{11}{c}{\textit{\color{gray} TIR modality}} \\ \hline
MiniCPM-V~\cite{MiniCPM-V} & arXiv 2024 & 43.62 & 40.83 & 43.92 & 39.1 & 39.27 & 46.24 & 39.65 & 47.01 & 41.78 \\
MiniGPT-v2~\cite{minigpt-v2} & arXiv 2023 & 34.47 & 31.5 & 33.45 & 28.65 & 33.05 & 33.34 & 28.44 & 35.91 & 34.45 \\
DeepSeek-VL~\cite{deepseek-vl} & arXiv 2024 & 39.2 & 34.63 & 38.39 & 33.27 & 34.46 & 39.09 & 36.34 & 39.71 & 33.93 \\
GeoPix~\cite{geopix} & GRSM 2025 & 42.73 & 39.7 & 42.03 & 37.45 & 41.36 & 46.29 & 43.13 & 45.99 & 44.13 \\
GeoChat~\cite{geochat} & CVPR 2024 & 43.08 & 40.42 & 45.72 & 39.85 & 41 & 47.77 & 45.81 & 42.85 & 42.35 \\
Falcon~\cite{falcon} & arXiv 2025 & 33.59 & 40.03 & 39.73 & 40.51 & 34.71 & 38.34 & 50.24 & 49.42 & 36.97 \\
Qwen2.5-VL-7B~\cite{qwen2.5-vl} & Ali 2025 & 40.57 & 38.98 & 40.49 & 38.41 & 37.89 & 44.12 & 46.76 & 43.43 & 41.9 \\
Qwen3-VL-8B (Base)~\cite{qwen3-vl} & Ali 2025 & 44.09 & 28.68 & 41.23 & 30.13 & 26.87 & 49.15 & 35.47 & 33.48 & 29.23 \\
GPT-4o~\cite{gpt4o} & OpenAI 2024 & 63.66 & 69.97 & 70.67 & 71.13 & 66.74 & 71.98 & 65.72 & 76.06 & 59.23 \\
Gemini-2.5-Flash~\cite{gemini2.5pro} & Google 2025 & 57.01 & 61.53 & 61.58 & 57.55 & 59.04 & 69.24 & 64.3 & 71.39 & 51.13 \\
Gemini-2.5-Pro~\cite{gemini2.5pro} & Google 2025 & 51.5 & 57.51 & 55.14 & 53.18 & 55.17 & 64.76 & 47.87 & 61.39 & 46.98 \\
MTCNet (Few-Shot) & - & 44.09 & 41.23 & 28.68 & 30.13 & 26.87 & 49.15 & 35.47 & 33.48 & 29.23 \\
\midrule
\multicolumn{11}{c}{\textit{\color{gray} MUL modality}} \\ \hline
MiniCPM-V~\cite{MiniCPM-V} & arXiv 2024 & 45.37 & 39.28 & 60.06 & 36.46 & 38.61 & 47.93 & 42.02 & 46.13 & 41.74 \\
MiniGPT-v2~\cite{minigpt-v2} & arXiv 2023 & 36.68 & 35.91 & 37.69 & 34.2 & 38.11 & 39.23 & 31.75 & 39.2 & 42.54 \\
DeepSeek-VL~\cite{deepseek-vl} & arXiv 2024 & 42.5 & 38.14 & 57.19 & 36.41 & 36.81 & 44.74 & 40.13 & 43.28 & 35.78 \\
GeoPix~\cite{geopix} & GRSM 2025 & 43.62 & 38.11 & 51.42 & 37.21 & 39.9 & 45.12 & 37.91 & 43.21 & 41.17 \\
GeoChat~\cite{geochat} & CVPR 2024 & 44.22 & 41.66 & 51.19 & 41.8 & 41.65 & 45.27 & 43.92 & 43.87 & 47.19 \\
Falcon~\cite{falcon} & arXiv 2025 & 34.8 & 38.39 & 36.83 & 37.05 & 33.02 & 38.62 & 42.97 & 46.06 & 37.89 \\
Qwen2.5-VL-7B~\cite{qwen2.5-vl} & Ali 2025 & 43.73 & 39.75 & 43.2 & 39.81 & 39.21 & 47.59 & 49.61 & 43.87 & 42.73 \\
Qwen3-VL-8B (Base)~\cite{qwen3-vl} & Ali 2025 & 47.62 & 35.64 & 47.47 & 36.13 & 32.54 & 54.84 & 45.3 & 43.56 & 36.45 \\
GPT-4o~\cite{gpt4o} & OpenAI 2024 & 67.72 & 75.28 & 73.96 & \underline{77.67} & 73.13 & 74.06 & 69.51 & 78.03 & 69.13 \\
Gemini-2.5-Flash~\cite{gemini2.5pro} & Google 2025 & 67.47 & 74.92 & 73.65 & 75.91 & 71.37 & \underline{75.39} & 75.99 & \underline{79.2} & \underline{69.76} \\
Gemini-2.5-Pro~\cite{gemini2.5pro} & Google 2025 & 61.41 & 70.05 & 64.93 & 69.38 & 66.03 & 68.64 & 68.72 & 72.99 & 62.15 \\
MTCNet (Few-Shot) & - & 61.94 & 55.74 & 61.28 & 57.66 & 52.41 & 62.48 & 55.13 & 58.48 & 56.99 \\
Qwen3-VL-8B (Finetuned) & Ali 2025 & \underline{79.6} & \underline{80.55} & \underline{75.3} & 76.75 & \textbf{75.65} & \underline{79.77} & \underline{76.5} & 71.25 & 69.66 \\
Qwen3-VL-8B + MTCNet & - & \textbf{83.16} & \textbf{84.81} & \textbf{80.58} & \textbf{80.26} & 68.19 & \textbf{83.76} & \textbf{77.78} & \textbf{80.58} & \textbf{74.88} \\
\bottomrule
\end{tabular}
\end{table*}

\subsection{Quantitative Comparison}
\label{sec:quantitative}

We benchmark the proposed MTCNet against two categories of baselines: (1) Zero-shot Open-source MLLMs (e.g., MiniCPM-V, GeoChat, Qwen3-VL Base); (2) Closed-source Commercial MLLMs (e.g., GPT-4o, Gemini-2.5-flash). The quantitative results, evaluated across single and multiple modalities, are summarized in Table~\ref{tab:traffic_vqa_full_acc}.

\subsubsection{Performance on Cognitive and Perceptual Tasks}

As shown in Table~\ref{tab:traffic_vqa_full_acc}, MTCNet achieves state-of-the-art performance across the board, with a consistent and substantial advantage. It is particularly strong in complex cognitive scenarios, effectively bridging the domain knowledge gap inherent in general-purpose models.

\myparagraph{Superiority in Cognitive Understanding.}
The most critical evaluation domain is Traffic Cognitive Questions, which require an implicit understanding of traffic regulations combined with advanced contextual interpretation. While the Qwen3-VL-8B (Finetuned) baseline achieves a commendable cognitive accuracy of 80.55\%---validating the utility of supervised domain adaptation---our fully integrated Qwen3-VL-8B + MTCNet architecture improves this to 84.81\%, a margin of +4.26\%. This improvement indicates that standard fine-tuning is insufficient to capture the multi-layered logic governing complex traffic violations. The advantage of MTCNet stems from the PGKE module, which retrieves expert prototypes from the external TRM, grounding the cognitive interpretation process in explicit, formalized domain knowledge rather than fragile statistical correlations.

Furthermore, MTCNet demonstrates strong performance in fine-grained violation detection, achieving notable accuracies in Vehicle Violation (80.26\%) and Pedestrian Violation (77.78\%). Notably, our method significantly outperforms leading commercial models such as GPT-4o (Cognitive: 75.28\%) and Gemini-2.5-flash (Cognitive: 74.92\%), demonstrating that a domain-specialized, knowledge-anchored module can substantially surpass large-scale general-purpose models in highly regulated vertical domains.

\myparagraph{Robustness in Perceptual Tasks.}
For Traffic Perceptual Questions (e.g., object localization, presence detection), the fully fine-tuned MTCNet attains 80.26\% accuracy, surpassing the corresponding fine-tuned baseline (75.30\%) by nearly 5\%. This improvement verifies that the QASC module effectively enhances feature distinctiveness, particularly for small and densely packed UAV targets. Even when constrained to the few-shot regime, our method (MUL OA: 61.94\%) substantially outperforms all evaluated zero-shot open-source baselines (peak OA around 47.62\%), validating that our lightweight residual injection strategy preserves general perception abilities while improving sensitivity to domain-specific semantic objects.

\subsubsection{Evaluation of Multi-Spectral Robustness}

To assess the effectiveness of the QASC module under varying environmental conditions, we analyze performance across different input modalities (OPT, TIR, and MUL) as shown in Table~\ref{tab:traffic_vqa_full_acc}.

A recurring limitation in zero-shot baselines is that multi-modal (MUL) performance often stagnates or degrades relative to single optical (OPT) performance. For instance, Qwen3-VL-8B (Base) shows only a marginal shift from an OPT Overall Accuracy of 47.56\% to a MUL OA of 47.62\%, and GeoChat experiences a similarly negligible gain (OPT: 44.00\% vs. MUL: 44.22\%). This pattern suggests that naive feature concatenation introduces detrimental noise, rendering complementary thermal data largely ineffective without explicit semantic alignment.

In contrast, MTCNet consistently achieves its best performance in the MUL setting. Under the few-shot protocol, MTCNet improves the OPT baseline (47.56\%) to a MUL OA of 61.94\% (+14.38\%). This result confirms that the QASC module successfully facilitates constructive interaction between modalities. By routing bidirectional attention toward the most informative regions of the TIR channel, the model effectively compensates for optical degradations (e.g., deep shadows, glare, or extreme low light) without incurring performance penalties from cross-modal feature conflict, ensuring reliable performance across all-weather UAV surveillance conditions.

\subsection{Ablation Studies}
\label{sec:ablation}

To better understand the contributions of each architectural component, we conduct ablation studies targeting the core modules and fusion strategies, all under the few-shot setting to highlight baseline structural differences.

\subsubsection{Effectiveness of Proposed Modules}
We investigate the individual and combined contributions of the PGKE and QASC modules by incrementally integrating them into the baseline architecture (Qwen3-VL with standard concatenation). The results are reported in Table~\ref{tab:module_ablation}.

\begin{table}[h]
\centering
\caption{Ablation study isolating the PGKE and QASC modules under the few-shot setting. Example 1 represents the unmodified baseline, where features are aggregated element-wise. \textbf{OA}: Overall Accuracy, \textbf{AA}: Average Accuracy.}
\label{tab:module_ablation}
\resizebox{0.95\columnwidth}{!}{
\begin{tabular}{c|cc|cc}
\toprule
\textbf{ID} & \textbf{PGKE (Cognition)} & \textbf{QASC (Perception)} & \textbf{OA (\%)} & \textbf{AA (\%)} \\
\midrule
Exp1 & \XSolidBrush & \XSolidBrush & 47.62 & 46.93 \\
Exp2 & \CheckmarkBold & \XSolidBrush & 60.67 & 59.41 \\
Exp3 & \XSolidBrush & \CheckmarkBold & 61.27 & 59.86 \\
Exp4 & \CheckmarkBold & \CheckmarkBold & \textbf{61.94} & \textbf{59.97} \\
\bottomrule
\end{tabular}
}
\end{table}

\myparagraph{Effectiveness of PGKE (Exp2).} Adding the PGKE module alone yields a substantial gain in Overall Accuracy (OA), rising from the baseline's 47.62\% to 60.67\%. This improvement highlights the vital role of explicitly injecting domain-specific knowledge. Without the external TRM, the baseline model struggles to map elementary visual features to abstract traffic rules. The PGKE module effectively bridges this semantic gap by grounding visual representations in expert regulatory prototypes.

\myparagraph{Effectiveness of QASC (Exp3).} Integrating only the QASC module also produces a significant performance improvement, achieving an OA of 61.27\%. This gain highlights the necessity of dynamic, bidirectional context exchange between the optical and thermal modalities. By selectively compensating for degraded features, the QASC module establishes a robust perceptual foundation, which is essential for reliably identifying small and densely distributed traffic objects under adverse conditions.

\myparagraph{Synergy of PGKE and QASC (Exp4).} The fully integrated framework achieves the best overall performance (OA 61.94\%, AA 59.97\%). The gain over single-module variants suggests that perceptual robustness and knowledge-guided reasoning are complementary. Specifically, QASC improves the stability of input features under changing environmental conditions, which in turn allows PGKE to retrieve more accurate and contextually relevant prototypes to support complex reasoning.

\subsubsection{Analysis of Fusion Mechanisms}
We further compare the proposed QASC module against standard fusion operations, as shown in Table~\ref{tab:fusion_ablation}.

\begin{table}[h]
\centering
\caption{Comparison of different modality integration methods. The proposed QASC module consistently outperforms all static fusion operations.}
\label{tab:fusion_ablation}
\resizebox{0.95\columnwidth}{!}{
\begin{tabular}{l|c|c}
\toprule
\textbf{Integration Method} & \textbf{Overall Accuracy (OA)} & \textbf{$\Delta$ vs. Baseline} \\
\midrule
Optical Only (Baseline) & 47.56 & - \\
Thermal Only & 44.09 & - \\
\midrule
Element-wise Add & 47.62 & +0.06 \\
Concatenation & 49.12 & +1.56 \\
\textbf{MTCNet (QASC)} & \textbf{61.94} & \textbf{+14.38} \\
\bottomrule
\end{tabular}
}
\end{table}

\begin{figure}
  \centering
  \includegraphics[width=0.95\columnwidth]{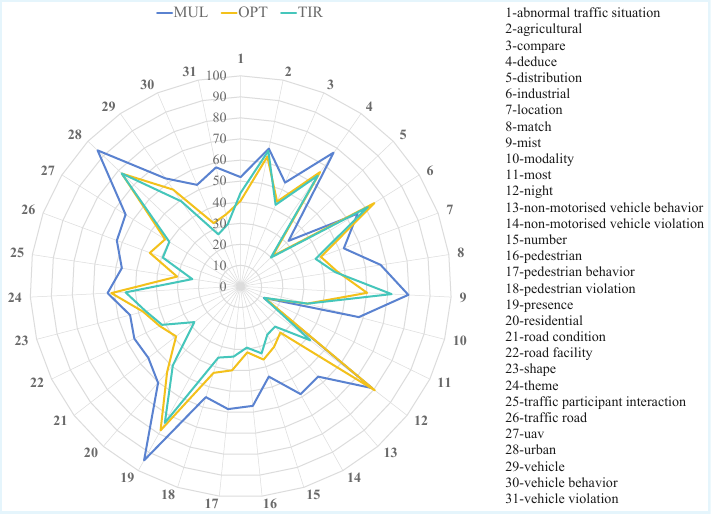}
  \caption{Performance comparison (quantified via CIDEr score) across the 31 question types in Traffic-VQA. The radar chart highlights the operational stability of the proposed multi-modal (MUL) approach, particularly in challenging environmental categories such as "Night" and "Mist," where single-modality approaches consistently underperform.}
  \label{fig:radar}
\end{figure}

Simple element-wise addition and concatenation yield OA values of 47.62\% and 49.12\%, respectively, offering only marginal improvements over the single-modality baseline. These results confirm that rigid, static fusion strategies fail to adaptively weight the importance of optical versus TIR features, a shortcoming that is exacerbated when one modality is heavily degraded by environmental noise. In contrast, our QASC mechanism achieves an OA of 61.94\%, demonstrating that dynamic, attention-based bidirectional context exchange is essential for robust all-weather traffic analysis.

\subsubsection{Fine-Grained Analysis by Question Type}
To evaluate the model's performance under specific environmental and task-oriented conditions, we conduct a fine-grained analysis structured across different question types, as visualized in the radar chart (Fig.~\ref{fig:radar}).

The breakdown reveals two key observations regarding environmental robustness:

\myparagraph{Night Scenarios.} For queries tagged with "Night", the optical-only (OPT) baseline suffers a significant performance drop (CIDEr: 12.93), while the standalone TIR modality remains robust (CIDEr: 80.55). Our fused model (MUL) effectively leverages the TIR stream to maintain strong performance (CIDEr: 78.84), demonstrating that the QASC module correctly identifies and prioritizes the reliable TIR channel when the optical stream is rendered ineffective by darkness.

\myparagraph{Fog or Mist Scenarios.} Under "Mist" conditions, the fused model (MUL) achieves a CIDEr score of 79.69, outperforming both single-modality inputs (OPT: 60.13, TIR: 71.75). This result indicates that even when both sensors are partially degraded by fog, the cross-spectral fusion strategy allows the model to aggregate complementary cues, yielding a result superior to either individual modality.

MTCNet shows consistently strong performance across a diverse set of tasks, spanning both high-level cognitive analysis (e.g., "Abnormal Traffic Situation") and low-level perception queries (e.g., "Location"). These results suggest that the model generalizes well across task types and is a practical candidate for real-world UAV traffic surveillance scenarios.

\section{Conclusion}

In this paper, we construct a large-scale, unified benchmark dataset, Traffic-VQA, designed to advance all-weather UAV traffic cognitive understanding. Comprising 8,180 well-aligned OPT-TIR image pairs and over 1.3 million question-answer pairs, the dataset covers diverse environmental conditions and 31 distinct cognitive tasks. To address the limitations of existing methods regarding domain knowledge deficiency and cross-modality interference, we propose MTCNet. The MTCNet incorporates a PGKE module working in conjunction with an external TRM to inject domain-specific situational knowledge into visual features. Furthermore, the QASC module adaptively integrates complementary spectral information through dynamic, attention-driven context exchange. Extensive experiments on Traffic-VQA demonstrate the effectiveness of MTCNet, which significantly outperforms contemporary state-of-the-art MLLMs, particularly in high-level cognitive scenarios.

We further highlight several directions for future research. The first concerns the transition from static spatial observation to continuous dynamic analysis. While Traffic-VQA provides a solid foundation for multi-spectral image comprehension, practical traffic surveillance inherently requires the analysis of continuous behavioral trajectories. Extending the benchmark toward video-based spatio-temporal VQA would enhance comprehensive event interpretation capabilities. The second direction focuses on the development of more powerful cognitive mechanisms. Compared to generalized visual perception tasks, specialized traffic behavior understanding relies heavily on regulatory constraints, highlighting the importance of integrating explicit prior knowledge into large-scale foundation models rather than relying solely on data-driven statistical correlations. In future work, we plan to extend this framework to video-based UAV-VQA and investigate the integration of additional complementary sensor modalities, with the goal of advancing robust, all-weather intelligent transportation systems.

\section*{Acknowledgement}
This work was supported in part by the National Natural Science Foundation of China (No. 62306005, 62006002, and 62076003), in part by the Natural Science Foundation of Anhui Province (No. 2208085J18 and 2208085QF192), and in part by the Natural Science Foundation of Anhui Higher Education Institution (No. 2022AH040014).

\bibliographystyle{unsrt}

\bibliography{mycankaowenxian}

\end{sloppypar}
\end{document}